\def\year{2021}\relax
\documentclass[letterpaper]{article} % DO NOT CHANGE THIS
\usepackage{aaai21}  % DO NOT CHANGE THIS
\usepackage{times}  % DO NOT CHANGE THIS
\usepackage{helvet} % DO NOT CHANGE THIS
\usepackage{courier}  % DO NOT CHANGE THIS
\usepackage[hyphens]{url}  % DO NOT CHANGE THIS
\usepackage{graphicx} % DO NOT CHANGE THIS
\def\UrlFont{\rm}  % DO NOT CHANGE THIS
\usepackage{natbib}  % DO NOT CHANGE THIS AND DO NOT ADD ANY OPTIONS TO IT
\usepackage{caption} % DO NOT CHANGE THIS AND DO NOT ADD ANY OPTIONS TO IT
\usepackage{amsmath}
\usepackage{commath}
\DeclareMathOperator*{\argmax}{argmax}
\usepackage{amsfonts}       % blackboard math symbols
\title{Learning Rewards from Linguistic Feedback}
\author {
    Theodore R. Sumers,\textsuperscript{\rm 1}
    Mark K. Ho,\textsuperscript{\rm 2}
    Robert D. Hawkins,\textsuperscript{\rm 2}\\
    Karthik Narasimhan,\textsuperscript{\rm 1}
    Thomas L. Griffiths\textsuperscript{\rm 1,2}\\
}
\affiliations {
    \textsuperscript{\rm 1}Department of Computer Science, Princeton University, Princeton, NJ \\
    \textsuperscript{\rm 2}Department of Psychology, Princeton University, Princeton, NJ \\
    \{sumers, mho, rdhawkins, karthikn, tomg\}@princeton.edu \\
}

\begin{document}

\maketitle

\begin{abstract}

We explore unconstrained natural language feedback as a learning signal for artificial agents. Humans use rich and varied language to teach, yet most prior work on interactive learning from language assumes a particular form of input (e.g., commands). We propose a general framework which does not make this assumption, instead using aspect-based sentiment analysis to decompose feedback into sentiment over the features of a Markov decision process. We then infer the teacher's reward function by regressing the sentiment on the features, an analogue of inverse reinforcement learning. To evaluate our approach, we first collect a corpus of teaching behavior in a cooperative task where both teacher and learner are human. We implement three artificial learners: sentiment-based ``literal'' and ``pragmatic'' models, and an inference network trained end-to-end to predict rewards. We then re-run our initial experiment, pairing human teachers with these artificial learners. All three models successfully learn from interactive human feedback. The inference network approaches the performance of the ``literal'' sentiment model, while the ``pragmatic'' model nears human performance. Our work provides insight into the information structure of naturalistic linguistic feedback as well as methods to leverage it for reinforcement learning.

\end{abstract}

\section{Introduction}
For autonomous agents to be widely usable, they must be responsive to human users' natural modes of communication. For instance, imagine designing a household cleaning robot. Some behaviors can be pre-programmed (e.g., how to use an outlet to recharge itself), while others must be learned (e.g., if a user wants it to charge in the living room or the kitchen). But how should the robot infer what a person wants? 

Here, we focus on \emph{unconstrained linguistic feedback} as a learning signal for autonomous agents. Humans use natural language flexibly to express their desires via commands, counterfactuals, encouragement, explicit preferences, or other forms of feedback. For example, if a human encounters the robot charging in the living room as desired, they  may provide feedback such as ``Great job.'' If they find it charging in the kitchen, the human may respond with ``You should have gone to the living room'' or ``I don't like seeing you in the kitchen.'' Our approach of learning rewards from such open-ended language differs from previous methods for interactive learning that use non-linguistic demonstrations~\cite{abbeel2004apprenticeship,argall2009survey,ho2016showing}, rewards/punishments~\cite{Knox2009Interactively, macglashan2017interactive, christiano2017deep}, or language commands~\cite{Tellex2011, Wang_2016, Tellex2020}. The agent's learning challenge is to interpret naturalistic feedback in the context of its behavior and environment to infer the teacher's preferences.

We formalize this inference as linear regression over features of a Markov decision process (MDP). We first decompose linguistic feedback into a scalar sentiment and a target subset of the MDP's features, a form of aspect-based sentiment analysis~\cite{hu2004mining, liu2020sentiment}. We then regress the sentiment against the features to infer the teacher's reward function. This enables learning rewards from arbitrary language. To extract target features, we first ground utterances to elements of the MDP~\cite{harnad1990symbol, mooney2008learning}. For example, ``Good job'' refers to prior behavior, whereas ``You should have gone to the living room'' refers to an action. This grounding determines the relevant MDP features: intuitively, positive sentiment about an action implies positive rewards on its features. We implement two versions of this model: a ``literal'' learner using only the explicit sentiment and a ``pragmatic'' learner with additional inductive biases~\cite{grice1975logic}. Because these models rely on domain-specific lexical groundings, we develop a parallel and potentially more scalable approach: training an inference network end-to-end to predict latent rewards from human-human interactions. In our live evaluation, all three models learn from human feedback. The inference network and ``literal'' sentiment models perform similarly, while the ``pragmatic'' model approaches human performance. We outline related work in Section~\ref{related_work_section}, then introduce our sentiment model in Section~\ref{overall_rewards_from_language_section}. Section~\ref{experiment_section} describes our task and experiment and Section~\ref{models_section} details our model implementations. Finally, Section~\ref{results_and_analysis_section} discusses results and Section~\ref{conclusion_section} concludes.\footnote{Code and data: github.com/tsumers/rewards.}

\begin{figure*}[t]
\begin{center}
\includegraphics[width=17.5cm]{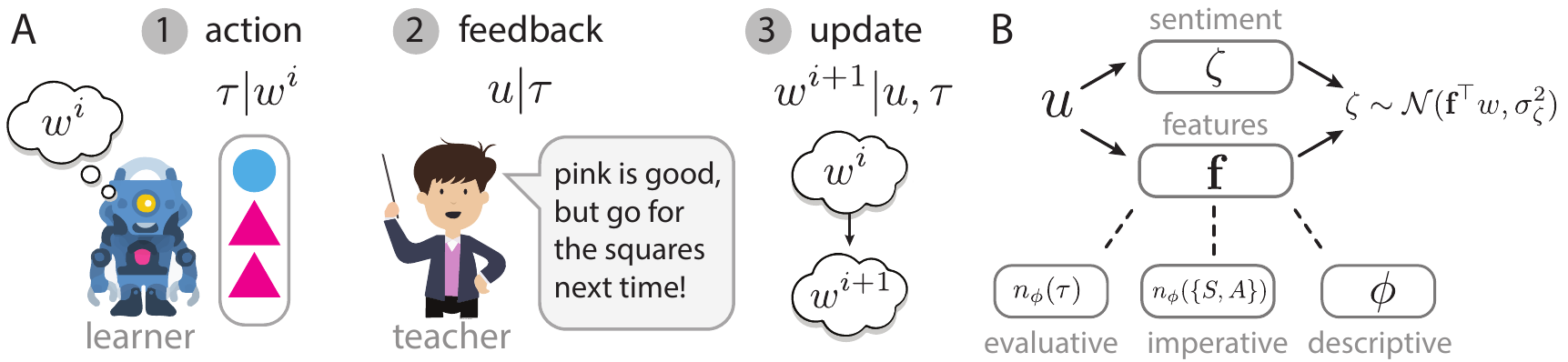}
\end{center}
\caption{A: Episodes involve three stages. B: We use aspect-based sentiment analysis to factor utterances into \textit{sentiment} and \textit{features}, then infer latent weights $w$ (solid lines). This allows us to integrate multiple forms of feedback (dashed lines).} 
\vspace{-1em}
\label{irl_language_schematic}
\end{figure*}

\section{Background and Related Work}
\label{related_work_section}
The work presented here complements existing methods that enable artificial agents to learn from and interact with humans. For example, a large literature studies how agents can learn latent preferences from non-linguistic human feedback. Algorithms such as TAMER~\cite{Knox2009Interactively} and COACH~\cite{macglashan2017interactive} transform human-generated rewards and punishments into quantities that reinforcement learning (RL) algorithms can reason with. Preference elicitation, which provides a user with binary choices between trajectories, is a similarly intuitive training method~\cite{christiano2017deep}. Finally, demonstration-based approaches use a set of expert trajectories to learn a policy---as in imitation learning~\cite{ross2010efficient}---or infer an underlying reward function---as in inverse reinforcement learning (IRL)~\cite{abbeel2004apprenticeship}. This idea has been extended to settings in which agents are provided with intentionally informative demonstrations~\cite{ho2016showing}, a variety of human acts~\cite{jeon2020rewardrational}, or themselves act informatively~\cite{Dragan2013Legibility, HadfieldMenell2016cooperative}. 

Another body of research has focused on linguistic human-agent interaction. Dialogue systems~\citep{artzi_2011_bootstrapping, li2016dialogue} learn to interpret user queries in the context of the ongoing interaction, while robots and assistants~\cite{thomason_2015, wang_2019_navigation, thomason_2020, szlam2019build} ground language in their physical surroundings. For a review of language and robotics, see~\citet{Tellex2020}. A parallel line of work in machine learning uses language to improve sample efficiency: to shape rewards \citep{maclin_1994, kuhlman_2004}, often via subgoals~\citep{kaplan_atari, tellex_2018_rewards, chevalier_boisvert_2018, goyal2019using, Bahdanau_2019, zhou2020inverse}. These approaches generally interpret and execute independent declarative statements (e.g., queries, commands, or (sub)goals). The most related approach to ours  performs IRL on linguistic input in the form of natural language commands~\citep{macglashan2015grounding, fu_2019_goals, goyal2020pixl2r}. 
Our work differs in two key ways: first, we use unconstrained and unfiltered natural language; second, we seek to learn general latent preferences rather than infer command-contextual rewards. A somewhat smaller body of work investigates such open-ended language: to correct captioning models \cite{Ling2017_captions}, capture environmental characteristics~\citep{narasimhan2018grounding}, or improve hindsight replay~\citep{cideron2019selfeducated}. For a review of language and RL, see~\citet{luketina2019survey}. 

We aim to recover the speaker's preferences from naturalistic interactions. Thus, unlike prior approaches, we do not \textit{solicit} a specific form of language (i.e. commands, corrections, or descriptions). We instead \textit{elicit} naturalistic human teaching and develop inferential machinery to learn from it. This follows studies of emergent language in other domains including action coordination~\citep{Djalali2011, Djalali_2012, potts_2012, ilinykh2019meetup, Suhr_2019}, reference pragmatics~\citep{He_2017, Udagawa_2019}, navigation~\citep{thomason2019visionanddialog}, and ``Wizard of Oz'' experiments~\citep{kim2009people, fraser2018_natural_language_agent}.

\section{Learning Rewards from Language}
\label{overall_rewards_from_language_section}
In this section, we formalize our approach. We develop a form of aspect-based sentiment analysis~\cite{hu2004mining, liu2020sentiment} to decompose utterances into sentiment and MDP features, then use linear regression to infer the teacher's rewards over those features. This allows us to perform an analogue of IRL on arbitrary language~\cite{abbeel2004apprenticeship}. To extract MDP features, we map utterances to elements within the teacher and learner's common ground~\cite{harnad1990symbol, clark1996using, mooney2008learning}, drawing on educational research to characterize typical communicative patterns~\citep{shute_2008, Lipnevich_education}.

\subsection{Setup}
We begin by defining a \emph{learner} agent whose interactions with the environment are defined by a Markov Decision Process (MDPs)~\cite{puterman1994markov}. 
Formally, a finite-horizon MDP $\mathcal{M} = \langle S, A, H, T, R \rangle$ is a set of states $S$, a set of actions $A$, a horizon $H \in \mathbb{N}$, a probabilistic transition function $T: S \times A \rightarrow \Delta(S)$, and a reward function $R: S \times A \rightarrow \mathbb{R}$. Given an MDP, a policy is a mapping from states to actions, $\pi: S \rightarrow A$.
An \emph{optimal policy}, $\pi^*$, is one that maximizes the future expected reward (value) from a state, $V^h(s) = \max_a R(s, a) + \sum_{s'} T(s' \mid s, a) V^{h - 1}(s')$, where $V^0(s) = \max_a R(s, a)$. 
States and actions are characterized by features $\phi$, where $\phi: S \times A \rightarrow \{0, 1\}^{K}$ is an indicator function representing whether a feature is present for a particular action $a$ in state $s$. 
We denote a state-action trajectory by $\tau = \langle s_0, a_0, ... s_T, a_T \rangle$. Finally, we define the feature counts over a set of state-action tuples as $n_\phi$: 
\begin{equation}
    n_\phi(\{\langle s,a\rangle\}) = \sum_{s,a\in\{\langle s,a\rangle\}}\phi(s,a)
\end{equation}
which we use to summarize sets of state-action tuples, including trajectories.

\subsection{Interactive Learning from Language}
\label{bayesian_lr_section}
We consider a setting where the reward function is hidden from the \emph{learner} agent but known to a \emph{teacher} agent who is allowed to send natural-language messages $u$ (Fig.~\ref{irl_language_schematic}A).
We formulate the online learning task as Bayesian inference over possible rewards: conditioning on the teacher's language and recursively updating a belief state.
Formally, we assume that the teacher's reward function is parameterized by a latent variable $w \in \mathbb{R}^K$ representing the rewards associated with features $\phi$:
\begin{equation}
    R(s,a) = w^{\top}\phi(s, a).
\end{equation}
We refer to these weights as the teacher's \textit{preferences} over features. 
The learner is attempting to recover the teacher's preferences from their utterances, calculating $P(w | u)$.

Learning unfolds over a series of interactive \textit{episodes}. 
At the start of episode $i$, the learner has a belief distribution over the teacher's reward weights, $P(w^{i})$, which it uses to identify its policy.
The learner first \emph{acts} in the world given this policy, sampling a trajectory $\tau^i$. 
They then receive \emph{feedback} in the form of a natural language utterance $u^i$ from the teacher (and optionally a reward signal from the environment). Finally, the learner uses the feedback to \emph{update} its beliefs about the reward, $P(w^{i+1} | u^i, \tau^i)$, which is then used for the next episode.

We now describe our general formal approach for inferring latent rewards from feedback. We first assume the learner extracts the \textit{sentiment} $\zeta$ and \textit{target features} $\mathbf{f}$ from the teacher's utterance, where $\mathbf{f} \in \mathbb{R}^K$ is a vector describing which features $\phi$ the utterance relates to. Extracting a sentiment and its target is known as \textit{aspect-based sentiment analysis}~\cite{liu2020sentiment}. Ready solutions exist to distill sentiment from  language~\cite{HuttoG14, Kim_2014}, but extracting the target features is more challenging. We detail our approach in Section~\ref{forms_of_feedback_section}. 

We then formalize learning as Bayesian linear regression:
\begin{equation}
    \zeta \sim \mathcal{N}(\mathbf{f}^{\top} w, \sigma_\zeta^2)
    \label{bayesian_lr_equation}
\end{equation}
We use a Gaussian prior: $w^i \sim \mathcal{N}(\mu_i, \Sigma_i)$. After each episode, we perform Bayesian updates~\cite{murphy_2007} to obtain a posterior: $P(w^{i+1} | u^i, \tau^i) = \mathcal{N}(\mu_{i+1},\Sigma_{i+1})$. Thus, similar to IRL methods~\citep{ramachandran2007bayesian}, the regression sets the teacher’s latent preferences $w \in \mathbb{R}^K$ to ``explain’' the sentiment. Intuitively, if the teacher says ``Good job,'' a learner could infer the teacher has positive weights on the features obtained by its prior trajectory. In the next section, we formalize this mapping to features. 

\subsection{Extracting MDP Features from Language}
\label{forms_of_feedback_section}
The main challenge of aspect-based sentiment analysis is extracting target features from arbitrary language. To accomplish this, we draw on educational research~\citep{Lipnevich_education, shute_2008}, which studies the characteristic forms of feedback given by human teachers. We first identify correspondences between these forms and prior work in RL. We then show each form targets a distinct element of the MDP (e.g., a prior trajectory). Mapping language to these elements allows us to extract target features.

\textbf{Evaluative Feedback}. Perhaps the simplest feedback an agent can receive is a scalar value in response to their actions (e.g., environmental rewards, praise, criticism). The RL literature has previously elicited such feedback (+1/-1) from human teachers~\citep{thomaz2008teachable, Knox2009Interactively, macglashan2017interactive}. In our setting, we consider how linguistic utterances can be interpreted as evaluative feedback. For example, ``Good job'' clearly targets the learner's behavior, $\tau^i$. We thus set the target features to the feature counts obtained by that trajectory: $\mathbf{f} = n_\phi(\tau^i)$.\footnote{As an example, an alternative teaching theory could use an inverse choice model~\citep{mcfadden1974conditional}. This would posit a teacher giving feedback on the learner's \textit{implied, latent} preferences, rather than their \textit{explicit, observed} actions.}

\textbf{Imperative Feedback}. Another form of feedback tells the learner what the correct action was. This is the general form of supervised learning. In RL, it includes labeling sets of actions as good or bad~\citep{judah2010_critique, christiano2017deep}, learning from demonstrations~\citep{ross2010efficient, abbeel2004apprenticeship, ho2016showing}, and corrections to dialogue agents~\citep{li2016dialogue, chen-etal-2017-line}. In our setting, imperative feedback specifies a counterfactual behavior: something the learner should (or should not) have done (e.g., ``You should have gone to the living room.''). Imperative feedback is thus a retrospective version of a command. Extracting features takes two steps: we first ground the language to a set of actions, then aggregate their feature counts. Formally, we define a state-action grounding function $G(u, S, A)$ which returns a set of state-action tuples from the full set: $G: u, S, A \mapsto \tilde{S}, \tilde{A}$, where $\tilde{S} \subseteq S, \tilde{A} \subseteq A$. We take the feature counts of these tuples as our target: $\textbf{f} = n_\phi(G(u, S, A))$.

\begin{figure*}[h!]
\begin{center}
\includegraphics[width=17.5cm]{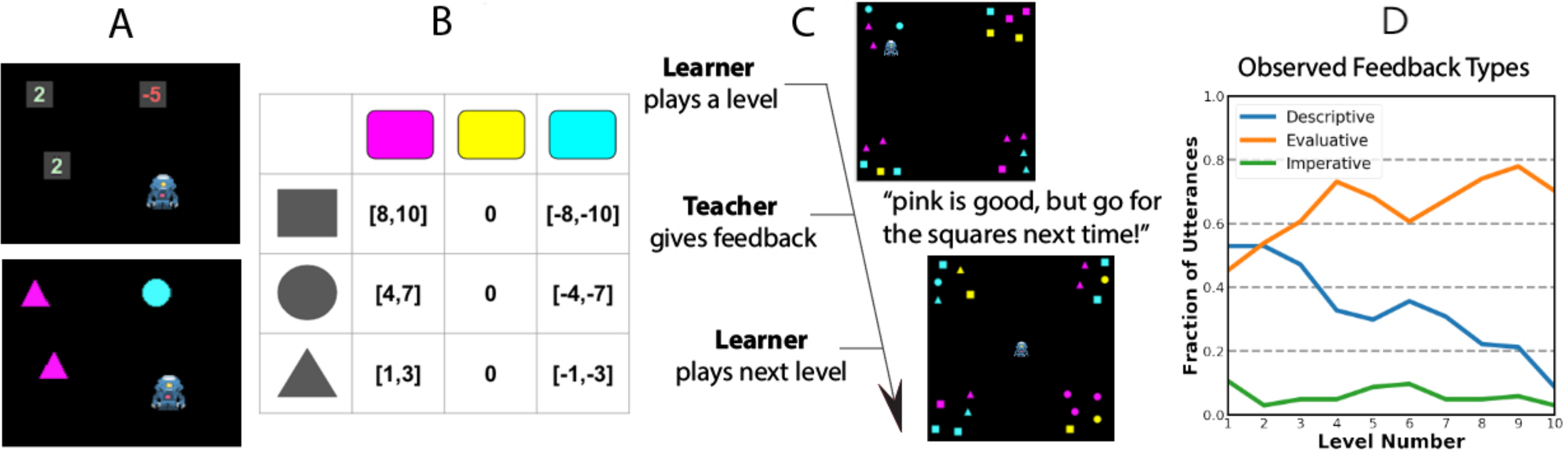}
% labeled_experiment_image_three_bars
\end{center}
\caption{A: The learner collected objects with different rewards (top) masked with shapes and colors (bottom). The teacher could see both views. B: Example reward function used to mask objects (here, objects worth 1-3 reward are rendered as pink triangles; the teacher thus ``prefers'' pink squares, worth 8-10). C: Pairs played 10 episodes, each on a new level. D: Feedback shifted from descriptive to evaluative as learners improved. Learners scored poorly on level 6, reversing this trend.} 
\label{task_ui}
\end{figure*}

\textbf{Descriptive Feedback}. Finally, descriptive feedback provides explicit information about how the learner should modify their behavior. Descriptive feedback is the most variable form of human teaching, encompassing explanations and problem-solving strategies. It is generally found to be the most effective~\citep{shute_2008, Lipnevich_education, van_der_klej_2015, Hattie_2007}. Supervised and RL approaches have used descriptive language to improve sample efficiency~\citep{srivastava2017_explanation, Hancock_2018_explanation, Ling2017_captions} or communicate general task-relevant information~\citep{narasimhan2018grounding}. In IRL, descriptive feedback explains the underlying structure of the teacher's preferences and thus relates directly to features $\phi$.\footnote{In problem settings beyond IRL, such feedback may relate to the transition function $T: S\times A$.} If the human says `I don't like seeing you in the kitchen,'' the robot should infer negative rewards for states and actions where it and the human are both in the kitchen. Formally, we define an indicator function over features designating whether or not that feature is referenced in the utterance: $I: u, \phi \rightarrow \{0,1\}^K$. We then set $\mathbf{f} = I(u, \phi)$.  

Prior RL algorithms generally operate on one of these forms. Interactions are constrained, as the algorithm \emph{solicits} feedback of a particular type. Our framework unifies them, allowing us to learn from a wide range of naturalistic human feedback. Concretely, we define a \textit{grounding} function $f_\text{G}:u \mapsto \textit{form}, \textit{ form} \in\{\text{Imperative, Evaluative, Descriptive}\}$, then extract $\mathbf{f}$ accordingly:
\begin{equation}
    \textbf{f} \in \mathbb{R}^K =  
\begin{cases}
    n_\phi(\tau^i) & \text{if } f_\text{G}(u) = \text{Evaluative} \\
   n_\phi(G(u, S, A)) & \text{if } f_\text{G}(u) = \text{Imperative} \\
    I(u, \phi) & \text{if } f_\text{G}(u) = \text{Descriptive} \\
\end{cases}
\label{forms_equation}
\end{equation}

This procedure is illustrated in Fig.~\ref{irl_language_schematic}. Table~\ref{tab:experiment_language_example} shows examples of this decomposition for various forms. In Section~\ref{experiment_section}, we elicit and analyze naturalistic human-human teaching, observing these three forms. In Section~\ref{models_section}, we describe a pair of agents implementing this model. Finally, we train an end-to-end neural network and probe its representations, showing that it learns to distinguish between forms directly from the data.

\section{Human-Human Instruction Dataset}
\label{experiment_section}

To study human linguistic feedback and generate a dataset to evaluate our method, we designed a two-player collaborative game (Fig.~\ref{task_ui}). One player (the learner) used a robot to collect a variety of colored shapes. Each yielded a reward, which the learner could not see. The second player (the teacher) watched the learner and could see the objects' rewards. Teacher-learner pairs engaged in 10 interactive episodes. We describe the experiment below.

\begin{table*}[h]
\begin{center} 
\vskip 0.12in
\begin{tabular}{lcccc}
\hline
Utterance   &  Feedback Form & Grounding ($f_G$) & Features (\textbf{f}) & Sentiment ($\zeta$)  \\
\hline
``Keep it up excellent'' & Evaluative & $n_\phi(\tau^i)$ & Behavior-dependent  & +17 \\
``Not a good move''  & Evaluative &  $n_\phi(\tau^i)$ & Behavior-dependent  & -10 \\
``Top left would have been better''  &  Imperative & $n_\phi(G(u, S, A))$ & Environment-dependent & +17 \\
``The light-blue squares are high valued'' & Descriptive & $I(u, \phi)$ & $\phi_{\text{BlueSquare}}$ & +13  \\
``I think Yellow is bad'' & Descriptive &  $I(u, \phi)$ & $\phi_\text{Yellow}$ & -16 \\
\hline
\end{tabular}
\caption{Example feedback from our experiment with feature / sentiment decomposition.}
\label{tab:experiment_language_example}
\end{center}
\end{table*}

\subsection{Experiment and Gameplay}
We recruited 208 participants from Amazon Mechanical Turk using psiTurk \cite{gureckis_2016}. Participants were paid \$1.50 and received a bonus up to \$1.00 based on the learner's score. The full experiment consisted of instructions and a practice level, followed by 10 levels of gameplay. Each level contained a different set of 20 objects. We generated 110 such levels, using 10 for the experiment and 100 for model evaluation (Section~\ref{results_and_analysis_section}). Collecting each object yields a reward between -10 and 10. Objects were distributed to each of the four corners (Fig.~\ref{task_ui}C). In each \emph{episode}, the learner had 8 seconds to \emph{act} (move around and collect objects), then the teacher had unlimited time to provide \emph{feedback} (send chat messages). Both players were shown the score and running bonus during the feedback phase. This leaked information about the reward function to the learner, but we found it was important to encourage active participation. The primary disadvantage is that the human baseline for human-model comparisons benefits from additional information not seen by our models.

\subsection{Task MDP and Rewards}
\label{task_mdp}
\label{reward_functions}
Human control was continuous: the learner used the arrow keys to steer the robot. However, the only rewarding actions were collecting objects and there was no discounting. As a result, the learner's score was the sum of the rewards of the collected objects. Due to object layout and short time horizon, learners only had time to reach one corner. Each corner had 5 objects, so there were 124 possible object combinations per level.\footnote{Choice of corner, then up to 5 objects: $4 * ({5 \choose 1} + {5 \choose 2} + {5 \choose 3} + {5 \choose 4} + {5 \choose 5}) = 124$. This can be seen as a 124-armed bandit.} We refer to these combinations as trajectories $\tau$, and formalize the task as choosing one. Concretely, the learner samples its beliefs $w \sim p(w^i)$, then chooses the optimal trajectory: $\pi := \argmax_\tau V^\tau = w^{\top} n_\phi(\tau)$. To induce teacher preferences, we assigned each teacher a reward function which masked objects with shapes and colors. Thus the distribution of actions and rewards on each level were the same for all players, but the objects were displayed differently depending on the assigned reward function. Our reward functions combined two independent perceptual dimensions, with color (pink, blue, or yellow) encoding sign and shape (circles, squares, or triangles) encoding magnitude (Fig.~\ref{task_ui}B). We permuted the shapes and colors to generate 36 different functions. 
    
\subsection{Human-Human Results and Language}
\label{observed_language_in_experiment}
Our 104 human pairs played 10 games each, yielding 1040 total messages (see Table~\ref{tab:experiment_language_example} for examples). We use our feedback classifier (see Section~\ref{literal_model_desc}) to explore the prevalence of various forms of feedback. We observe that humans use all three in a curriculum structure known as ``scaffolding''~\cite{shute_2008}: teachers initially use descriptive feedback to correct specific behavior, then switch to evaluative as the learners' score improves (Fig~\ref{task_ui}D). This can be seen as starting with ``off-policy'' feedback, then switching to ``on-policy'' evaluation. Teachers could send unlimited messages and thus sometimes used multiple forms. Most episodes contained \textit{evaluative} (63\%) or \textit{descriptive}  (34\%) feedback; fewer used \textit{imperative} (6\%). The infrequency of imperative feedback is reasonable given our task: specifying the optimal trajectory via language is more challenging than describing desirable features. Not all pairs fared well: some learners did not listen, leading teachers to express frustration; some teachers did not understand the task or sent irrelevant messages. We do not filter these out, as they represent naturalistic human language productions under this setting. 

\section{Agent Models}
\label{models_section}
\label{stateful_beliefs}
We now describe our three models. The first (Section~\ref{literal_model_desc}) directly implements our sentiment-based framework. The second (Section~\ref{pragmatic_model_desc}) extends it with pragmatic biases based on Gricean maxims~\citep{grice1975logic}. Finally, we train a neural net end-to-end from experiment episodes (Section~\ref{learned_model_desc}).

\subsection{``Literal'' Model}
\label{literal_model_desc}
\label{phrase_classifier}
\label{reference_types_details}
Our literal model uses a supervised classifier to implement $f_\text{G}$ and a small lexicon to extract target features. 

\textbf{Utterance Segmentation and Sentiment.} Teachers often sent multiple messages per episode, each potentially containing multiple forms of feedback. To process them, we first split each message on punctuation (!.,;), then treated each split from each message as a separate utterance. To extract sentiment, we used VADER \citep{HuttoG14}, which is optimized for social media. VADER provides an output $\zeta \in [-1,1]$, which we scaled by 30 (set via grid search). 

\textbf{Utterance Features.} To implement $f_\text{G}$, we labeled 685 utterances from pilot experiments and trained a logistic regression on TF-IDF unigrams and bigrams, achieving a weighted-F1 of .86. For evaluative feedback, as described in Eq.~\ref{forms_equation}, we simply used the feature counts from the learner's trajectory $\mathbf{f} = n_\phi(\tau^i)$. Imperative feedback requires a task-specific action-grounding function $G(u,S,A)$. While action grounding in complex domains is an open research area~\citep{Tellex2020}, in our experiment all imperative language referenced a cluster of objects (e.g. ``Top left would have been better''). We thus used regular expressions to identify references to corners and aggregated features over actions in that corner. For descriptive feedback, we defined a similar indicator function $I(u, \phi)$ identifying features in the utterance. We used relatively nameable shapes and colors, so teachers used predictable language to refer to object features (``pink'', ``magenta'', ``purple'', ``violet''...). Again, we used regular expressions to match these synonyms. Finally, we normalized $\norm{\mathbf{f}}_1 = 1$ so all forms carry equal weight. 

\textbf{Belief Updates}. Because players had seen object values in practice levels ranging between -10 and 10, we initialized our belief state as $\mu_0 = 0$, $\Sigma_0 = \mbox{\rm diag}(25)$. This gives an approximately 95\% chance of feature weights falling into that range. For each utterance, we perform Bayesian updates to obtain posteriors $P(w^{i+1} | u^i, \tau^i) = \mathcal{N}(\mu_{i+1},\Sigma_{i+1})$. We use $\sigma_\zeta^2 = \frac{1}{2}$ for all updates, which we set via grid search.

\subsection{``Pragmatic'' Model}
\label{pragmatic_model_desc}
We augment the ``literal'' model with two biases based on pragmatic principles \cite{grice1975logic}. While pragmatics are often derived as a result of recursive reasoning~\cite{goodman_2016}, we opt for a simpler heuristic approach.

\textbf{Sentiment Pragmatics.} The Gricean ``maxim of quantity'' states that speakers bias towards parsimony. Empirically, teachers often referenced a feature or an action without an explicit sentiment. Utterances such as ``top left'' or ``pink circles'' implied positive sentiment (e.g. ``pink circles [are good]''). To account for this, we defaulted to a positive bias ($\zeta = 15$) if the detected sentiment was neutral.  

\textbf{Reference Pragmatics.} The Gricean ``maxim of relation'' posits that speakers provide information that is relevant to the task at hand. We assume utterances describe important features, and thus unmentioned features are \textit{not} useful for decision making. We implemented this bias by following each Bayesian update with a second, negative update ($\zeta = -30$, set via grid search) to all features not referenced by the original update, gradually decaying weights of unmentioned features.

\subsection{End-to-end Inference Network}
\label{learned_model_desc}
To complement our lexicon-based sentiment models, we train a small inference network to predict the teacher's latent rewards. We use human data from our experiment to learn an end-to-end mapping from the $(u, \tau)$ tuples to the teacher's reward parameters. Conceptually, this is akin to ``factory-training'' a housecleaning robot, enabling it to subsequently adapt to its owners' particular preferences. 

\textbf{Model Architecture.} We use a feed-forward architecture. We tokenize the utterance and generate a small embedding space (D=30), representing phrases as a mean bag-of-words (MBOW) across tokens. We represent the trajectory with its feature counts, $n_\phi(\tau)$. We concatenate the token embeddings with the feature counts, use a single fully-connected 128-width hidden layer with ReLU activations, then use a linear layer to map down to a 9-dimension output.

\textbf{Model Training.} Our dataset is skewed towards positive-scoring games, as players learned over the course of the experiment. To avoid learning a default positive bias, we first downsample positive-score games to match negative-score ones. This left a total of 388 episodes from 98 different teachers with a mean score of 1.09 (mean of all games was 8.53). We augment the data by exchanging the reward function (Fig.~\ref{task_ui}B), simulating the same episode under a different set of preferences. We take a new reward function and switch both feature counts and token synonyms, preserving the relationships between $u^i, \tau^i$, and $w$. We repeat this for all 36 possible reward functions, increasing our data volume and allowing us to separate rewards from teachers. We used ten-fold CV with 8-1-1 train-validate-test splits, splitting both teachers and reward functions. Thus the network is trained on one set of rewards (i.e. latent human preferences) and teachers (i.e. linguistic expression of those preferences), then tested against unseen preferences and language. We used stochastic gradient descent with a learning rate of .005 and weight decay of 0.0001, stopping when validation set error increased. We train the network, including embeddings, end-to-end with an L2 loss on the true reward.   

\begin{figure}[t]
\begin{center}
\includegraphics[width=8.4cm]{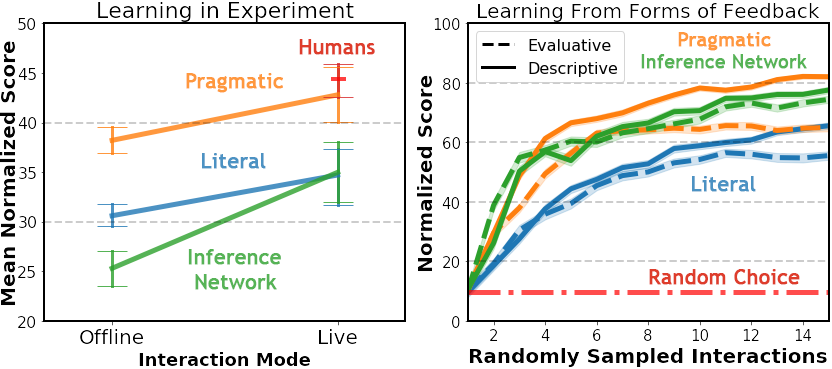}
\end{center}
\caption{Left: learning within our experiment structure. We plot averaged normalized score over the 10 learning episodes; bars indicate 1 SE (68\% CI). Right: learning with specific feedback types. We plot averaged normalized score on 100 test levels after each episode.}
\label{model_results_figure}
\end{figure}

\begin{figure*}[h]
\begin{center}
\includegraphics[width=14cm]{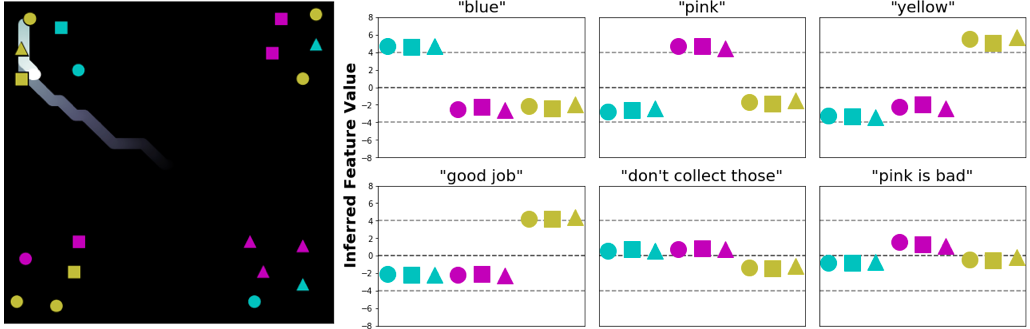}
\end{center}
\caption{Left: A trajectory from our experiment. Right: Inference network output given this trajectory. Top row: the model learns to map feature-related tokens (``Descriptive'' feedback) directly to rewards, independent of the trajectory. Bottom left / center: the model maps praise and criticism (``Evaluative'' feedback) through the feature-counts from the trajectory. Bottom right: a failure mode. ``Descriptive'' feedback with negative sentiment, a rare speech pattern, is not handled correctly.} 
\label{learned_utterance_targets}
\end{figure*}

\textbf{Multiple Episodes.} Given a $(u, \tau)$ tuple, our model predicts the reward $\hat{w}$ associated with every feature. To evaluate it over multiple trials in Section~\ref{results_and_analysis_section}, we use a comparable update procedure as our structured models. Concretely, we initialize univariate Gaussian priors over each feature $\mu_0 = 0$, $\sigma_0 = 25$, then run our inference network on each interaction and perform a Bayesian update on each feature using our model's output as an observation with the same fixed noise. For each feature, $P(w^{i+1} | u^i, \tau^i) = \mathcal{N}(\mu_{i},\sigma_{i}) * \mathcal{N}(\hat{w}, \frac{1}{2})$. In all offline testing, we use the network from the appropriate CV fold to ensure it is always evaluated on its test set (teachers and rewards).

\section{Results and Analysis}
\label{results_and_analysis_section}
We seek to answer several questions about our models. First, do they work: can they recover a human's reward function? Second, does our sentiment approach provide an advantage over the end-to-end learned model? And finally, do the ``pragmatic'' augmentations improve the ``literal'' model? We run a second interactive experiment pairing human teachers with our models and find the answer to all three is yes (Section~\ref{dyad_trajectory_comparison}). We then analyze how our models learn by testing forms of feedback separately (Section~\ref{random_sample_interactions}).

\tabcolsep=1.2mm
\begin{table}[t]
\begin{center} 
\begin{tabular}{l|lll|llll}
\textbf{Model}     & \multicolumn{3}{c|}{\textbf{Experiment}} & \multicolumn{4}{c}{\textbf{Interaction Sampling}} \\
                        & Offline  & Live  & n      &All            &Eval     &Desc        &Imp\\ \hline
\textbf{Literal}        &  30.6     & 34.7      & 46           & 40.5           & 38.7          & 40.6              & 16.7          \\
\textbf{Pragmatic}      & \textbf{38.2}      & \textbf{42.8}      & 47               & \textbf{52.5} & 50.4          & \textbf{58.2}   & \textbf{31.7}\\
\textbf{Inference}  & 25.3      & 35.0     & 55               & 47.6            & \textbf{54.3} & 53.2              & --         \\
\textbf{Human}          & --        & 44.3      & 104         & --                &--             &--             &--        
\end{tabular}
\caption{Normalized scores averaged over 10 episodes of learning. ``Experiment'' plays the 10 experiment episodes with a single human; ``Interaction Sampling'' draws ($u,\tau)$ tuples from the entire corpus and plays 100 test levels after each update.} 
\label{tab:pragmatics_and_utterance_type_table}
\end{center} 
\end{table}

\subsection{Learning from Live Interactions}
\label{dyad_trajectory_comparison}

To evaluate our models with live humans, we recruited 148 additional participants from Prolific, an online participant recruitment tool, and paired each with one of three model learners in our task environment. We measured the averaged normalized score across all 10 levels (the mean percentage of the highest possible score achieved). 
To assess the effect of interactivity, we also evaluated the same three model learners on replayed sequences of $(u,\tau)$ tuples from our earlier human-human experiment. The results are shown in Fig.~\ref{model_results_figure} and summarized in Table~\ref{tab:pragmatics_and_utterance_type_table}. 
We conducted a mixed-effects regression~\cite{kuznetsova2017lmertest} using performance as the dependent variable, including fixed effects of time (i.e. episode 1, episode 2, etc.), interactivity (i.e. live vs. offline), and learner model (i.e. neural vs. ``literal'' vs. ``pragmatic''), as well as an interaction between interactivity and time. 
We also included random intercepts and random effects for the learner model for each pair to control for clustered variance. 
The categorical factor of the learner model was contrast-coded to first compare the neural against the two sentiment models and then compare the two sentiment models directly against each other.

First, we found a significant main effect of time, $t(4138)=32.77, p<.001$, indicating that performance improves over successive levels. 
Second, although there was no significant main effect of interactivity, $t(446)=-.08, p=.94$, there was a significant interaction between  interactivity and time, $t(4138)=2.32, p=.02$, suggesting that the benefits of the live condition manifest over successive episodes as the teacher provides feedback conditioned on the learner's behaviors. 
Finally, turning to the models themselves, we find that the ``family'' of sentiment models collectively outperform the neural network $t(132)=3.57, p<.001$ and the ``pragmatic'' sentiment model outperforms the ``literal'' one, $t(147)=-2.37, p=.019$.
Post-hoc pairwise tests~\cite{tukey1953section} find an estimated difference of $d = 7.8, 95\%$ CI: $[2.7, 12.9]$ between the ``pragmatic'' and ``literal'' models; $d = -3.77, 95\%$ CI: $[-11.8, 4.3]$ between the neural and ``literal''; and $d = -11.5, 95\%$ CI: $[-19.4, -3.7]$ between neural and ``pragmatic.'' 
This suggests the end-to-end model learns to use most of the literal information in the data, while the inductive biases we encoded into the ``pragmatic'' model capture additional implicatures. 
\subsection{Learning from Different Forms of Feedback}
\label{random_sample_interactions}
\label{feedback_types}
To characterize model learning from different ``forms'' of feedback, we design a second evaluation independent of the experiment structure. Our ``episode'' sequence is as follows: we draw a $(u, \tau)$ tuple at random from the human-human experiment, \emph{update} each model, and have it \emph{act} on our 100 pre-generated test levels. We take its averaged normalized score on these levels. We repeat this procedure 5 times for each cross-validation fold, ensuring the learned model is always tested on its hold-out teachers and rewards. This draws feedback from a variety of teachers and tests learners on a variety of level configurations, giving a picture of overall learning trends. Normalized scores over test levels are shown in Fig.~\ref{model_results_figure} and Table~\ref{tab:pragmatics_and_utterance_type_table} (``Interaction Sampling''). All models improve when learning from the entire corpus (``All'') versus individual teachers (``Experiment''). The inference network improves most dramatically, suggesting it may be vulnerable to idiosyncratic communication styles used by individual teachers. We then use our feedback classifier (Section~\ref{literal_model_desc}) to expose models to only a single form of feedback. This reveals that our ``pragmatic'' augmentations help most on ``Descriptive'' feedback, which is critical for early learning in the experiment. Finally, we explore our inference network's contextualization process (Fig.~\ref{learned_utterance_targets}). It learns to map ``Evaluative'' feedback through its prior behavior and typical ``Descriptive'' tokens directly to the appropriate features. We also confirm failure modes on rarer speech patterns, most notably descriptive feedback with negative sentiment. This suggests the learned approach would benefit from more data.

\section{Conclusion}
\label{conclusion_section}
We presented two methods to recover latent rewards from naturalistic language: using aspect-based sentiment analysis and learning an end-to-end mapping from utterances and context to rewards. We find that three implementations of these models all learn from live human interactions. The ``pragmatic'' model in particular achieves near-human performance, highlighting the role of implicature in natural language. We also note that the inference network's performance varies qualitatively across evaluation modes: it outperforms the ``literal'' model when tested on the whole corpus, but ties it when playing with individual humans (``Interaction Sampling'' vs ``Experiment - Live''). This underscores the importance of evaluation in realistic interaction settings.

We see several future research directions. First, our sentiment models could be improved via theory-of-mind based pragmatics, while our end-to-end approach could benefit from stronger language models (recurrent networks or pre-trained embeddings). Hybridizing sentiment and learned approaches~\cite{jiang2011target, xu2019bertsentiment} could offer the best of both. We also see potential synergies with instruction following: treating commands as ``Imperative'' feedback could provide a preference-based prior for interpreting future instructions. Finally, we anticipate extending our approach to more complex MDPs in which humans teach both rewards and transition dynamics~\cite{narasimhan2018grounding}. In general, we hope the methods and insights presented here facilitate the adoption of truly natural language as an input for learning. 

\newpage

\section*{Acknowledgements}
We thank our anonymous reviewers for their thoughtful and constructive feedback. This work was supported by NSF grants \#1545126 and \#1911835, and grant \#61454 from the John Templeton Foundation. 

\section*{Ethics Statement}

Equipping artificial agents with the capacity to learn from linguistic feedback is an important step towards value alignment between humans and machines, with the end goal of supporting beneficial interactions. However, one risk is expanding the set of roles that such agents can play to those requiring significant interaction with humans --  roles currently restricted to human agents. As a consequence, certain jobs may be more readily replaced by artificial agents. On the other hand, being able to provide verbal feedback to such agents could expand the group of people able to interact with them, creating new opportunities for people with disabilities or less formal training in computer science.

\small 
\bibliography{references.bib}

\end{document}